\documentclass[10pt,twocolumn,letterpaper]{article}

\usepackage{cvpr}
\usepackage{times}
\usepackage{epsfig}
\usepackage{graphicx}
\usepackage{amsmath}
\usepackage{amssymb}
\usepackage{adjustbox}
\usepackage{booktabs}
\usepackage{threeparttable}
\usepackage{courier}
\usepackage{microtype}

\usepackage[breaklinks=true,bookmarks=false]{hyperref}

\cvprfinalcopy 

\def\cvprPaperID{3231} 

\begin{document}

\title{Saliency Weighted Convolutional Features for Instance Search}

\author{
Eva Mohedano$^1$~~
Kevin McGuinness$^2$~~
Xavier Gir\'o-i-Nieto$^1$~~
Noel E. O'Connor$^2$ \vspace{0.45em}
\\ 
$^1$ Insight Centre for Data Analytics, Dublin City University \\
$^2$ Universitat Polit\`{e}cnica de Catalunya \vspace{0.15em}\\
\footnotesize\tt \{eva.mohedano, kevin.mcguinness, noel.oconnor\}@insight-centre.org, \ xavi.giro@upc.edu}

\maketitle

\begin{abstract}

This work explores attention models to weight the contribution of local convolutional representations for the instance search task. We present a retrieval framework based on bags of local convolutional features (BLCF) that benefits from saliency weighting to build an efficient image representation. The use of human visual attention models (saliency) allows significant improvements in retrieval performance without the need to conduct region analysis or spatial verification, and without requiring any feature fine tuning. We investigate the impact of different saliency models, finding that higher performance on saliency benchmarks does not necessarily equate to improved performance when used in instance search tasks. The proposed approach outperforms the state-of-the-art on the challenging INSTRE benchmark by a large margin, and provides similar performance on the Oxford and Paris benchmarks compared to more complex methods that use off-the-shelf representations.

\end{abstract}

\section{Introduction}

Content-based image retrieval has been mostly tackled as the problem of instance-level image retrieval~\cite{Oxford5k,tembedding,vlad}. The notion of an ``instance'' specifically limits the search to ``one'' instance of a semantic class (i.e to retrieve instances of one specific object, person or location as in~\cite{Awad2017}). This contrasts with less specific retrieval pipelines in which any instance of any member of a class will satisfy the search.

\begin{figure}[t!]
\begin{center}
\includegraphics[width=0.48\textwidth]{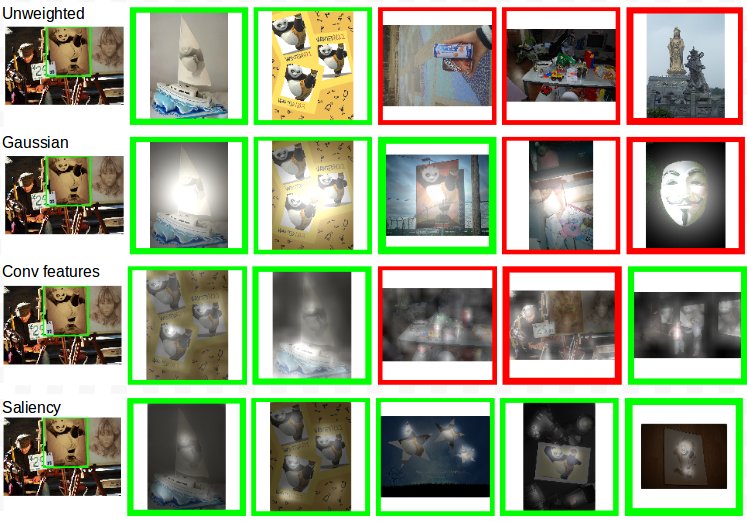}
\caption{Top 5 retrieved images for a query in INSTRE dataset using different spatial weighting schemes with BLCF. Correct results are highlighted in green, irrelevant images in red.}
\label{fig:saliency_models}
\end{center}
\end{figure}

Retrieval methods based on convolutional neural networks (CNN) have become increasingly popular in recent years. Off-the-shelf CNN representations extracted from convolution layers can be directly aggregated via spatial max/sum-pooling generating a global image representation that achieves a good precision and recall performance for a reasonable footprint~\cite{BabenkoL15,visualInstanceRazavianSMC14}. Moreover, when suitable training data is available, CNN models can be trained for similarity learning to adapt the learned representations to the end task of retrieval as well as to the target image domain (e.g. landmarks as in~\cite{gordo2016deep,Radenović2016}). However, the fine-tuning process involves the collection, annotation, and cleaning of a suitable training dataset, which is not always feasible. Furthermore, in generic instance search scenarios the target instances are unknown, which potentially makes off-the-shelf CNN representations more adequate for the task.

In this work we propose using state-of-the-art saliency models to weight the contribution of local convolutional features prior to aggregation. The notion of saliency in recent CNN-based image retrieval works has referred to the most active regions obtained from a specific filter within a CNN such as in~\cite{CaoLWHSS16,Jimenez_2017_BMVC}, or to a 2D feature saliency map derived from the convolutional layers~\cite{QAwrmac17,CroW,simeoni:unsupervisedObject}. In these cases, salient regions are detected with the same CNN from which image representations are extracted. On the other hand, human-based saliency models, such as those derived from the popular Itti and Koch model~\cite{itti1998model}, have been applied only in image retrieval pipelines based in local handcrafted features~\cite{Wu2017}. To the best of our knowledge, it has not yet been explored how the most recent CNN-based retrieval pipelines can benefit from modern saliency models, and how different models affect the retrieval performance. 

The contributions of this paper are the following:
\begin{itemize}

\item We propose a novel approach to instance search combining saliency weighting over off-the-shelf convolutional features which are aggregated using a large vocabulary with a bag of words model. The bags of local convolutional features (BLCF) scheme has the advantage of retaining the sparsity and scalability of the original bag of words model while giving improved performance across multiple benchmarks.

\item We demonstrate that this weighting scheme gives outperforms all other state of the art on the challenging INSTRE benchmark, without requiring any feature fine tuning.
Furthermore, it also offers comparable or better performance to more complicated encoding schemes for off-the-shelf features on Oxford and Paris benchmarks. 

\item We investigate the impact of the specific visual attention model used, and show that, perhaps surprisingly, higher performance on saliency benchmarks does not necessary equate to improved performance when used in the instance search task.
\end{itemize}

The remainder of the paper is organized as follows: Section~\ref{sec:related_work} presents recent CNN-based retrieval methods, Section~\ref{sec:BLCF} describes the proposed retrieval pipeline, Section~\ref{sec:saliency} introduces the different saliency models evaluated, and finally, Section~\ref{sec:experiments} presents the datasets and retrieval results obtained.

\section{Related Work}
\label{sec:related_work}

Several works have investigated the usage of convolutional layers for retrieval. Early approaches focused on the usage of fully connected layers as global image representation~\cite{neuralcodes}, that when combined with spatial search and data augmentation strategies~\cite{cnnofftheshelf} achieved a significant boost in performance in several retrieval benchmarks. This improvement came at the cost of significantly increasing the memory and computational requirements, meaning that the the system was not scalable to larger more realistic scenarios.


A second generation of works explored the convolutional layers to derive image representations for retrieval. In these approaches, the activations of the different neuron arrays across all feature maps in a convolutional layer are treated as local features. Babenko~\cite{BabenkoL15} have shown that a simple spatial sum pooling on the convolutional layer outperformed fully connected layers using pre-trained CNN models for retrieval for a reasonable footprint of 512D. Furthermore, the power of this representation could be enhanced by applying a simple Gaussian center prior scheme to spatially weight the contribution of the activations on a convolutional layer prior to aggregation. Following a similar line, Kalantidis~\cite{CroW} proposed a cross dimensional weighting scheme (CroW) that consisted of the channel sparsity of a convolutional layer and a spatial weighting based in the $L^2$-norm of the local convolutional features. The spatial weighting scheme directly depended on the image content, which was shown to be a more effective approach than the fixed center prior approach. 

A different line are the works performing region analysis using convolutional features. Tolias \textit{et al.}~\cite{rmac-ToliasSJ15} proposed the popular regional of maximum activation (R-MAC). This is an aggregation method for convolutional features that, given a convolutional layer, generates a set of regional vectors by performing spatial max-pooling within a particular region. The regions are individually post-processed (by $L^2$-normalization, followed by a PCA whitening, and a second $L^2$-normalization), and aggregated via sum pooling into a single compact representation. One of the main limitations of the approach is using a fixed grid of locations. Several works has been proposed to overcome the R-MAC limitations. Jimenez \textit{et al.}~\cite{Jimenez_2017_BMVC} explored the Class Activation Maps (CAMs) to improve over the fix region strategy of R-MAC. CAMs generates a set of spatial maps highlighting the contribution of the areas within an image that are most relevant to classify an image as a particular class. In their approach, each of the maps is used to weight the convolutional features and generate a set class vectors, which are post-processed and aggregated as the region vectors in R-MAC. Similarly, Cao \textit{et al.}~\cite{CaoLWHSS16} proposed a method to derive a set of base regions directly from the convolutional layer, followed with a query adaptive re-ranking strategy. Lanskar and Kannala~\cite{QAwrmac17} used a saliency measure directly derived from the convolutional features to weight the contribution of the regions of R-MAC prior to aggregation. Their proposed saliency measure consists in sum aggregating the feature maps over the channel dimension. Similarly, the very recent work of Simeoni \textit{et al} ~\cite{simeoni:unsupervisedObject} propose using a saliency measure also derived from a convolutional layer that consists in a weighted sum across channels where weights represent the channel sparsity. They propose a method to detect a set of rectangular regions directly from the feature saliency map, improving over the uniform sampling of R-MAC.

The current state-of-the-art retrieval approaches~\cite{gordo2016deep,Radenović2016} use models fine-tuned with a ranking loss. In particular, Gordo~\emph{et al.}~\cite{gordo2016deep} improves over the original R-MAC encoding by learning a region proposal network~\cite{FastERGirshick15} for the task of landmark retrieval. Region proposal and feature learning are optimized end-to-end  achieving excellent performance in the popular Paris, Oxford, and Holidays datasets. This approach, however, requires the construction of a suitable training dataset, which is usually domain specific,  time consuming to construct and it is unclear how those models generalize in more generic scenarios such as the INSTRE dataset.

\begin{figure}[t!]
\begin{center}
\includegraphics[width=.48\textwidth]{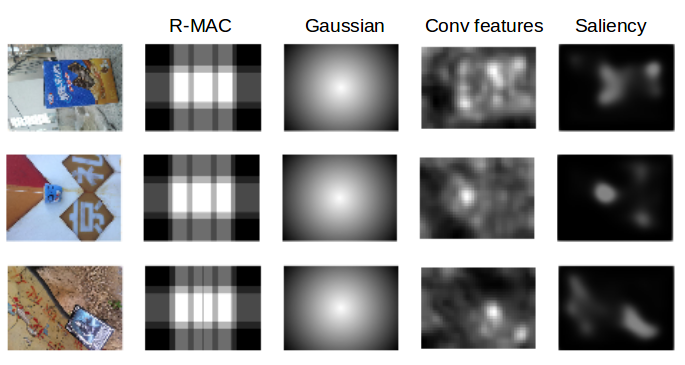}
\caption{Spatial weighting schemes used in the literature on three INSTRE images. R-MAC visualizes the window location strategy used in~\cite{rmac-ToliasSJ15}, Gaussian center prior is used in~\cite{BabenkoL15}, saliency measure derived from the local convolutional features used in~\cite{CroW,QAwrmac17,simeoni:unsupervisedObject} (we visualize the $L^2$-norm of each local feature), and human-based saliency prediction used in this work from SalGAN model~\cite{SalGAN}.}
\label{fig:weighting_schemes}
\end{center}
\end{figure}

Intuitively, one can see that pooling descriptors from different regions and then subsequently pooling all of the resulting together is similar to applying a weighting scheme to the original convolutional features since region descriptors and globally pooled descriptors are built from the same set of local features. Figure~\ref{fig:weighting_schemes} visualizes the region sampling of the original R-MAC~\cite{rmac-ToliasSJ15} as a spatial weighting strategy (without considering the region post-processing). The approach provides more emphasis to the center located CNN features. In contrast to the approaches in the literature, we propose a simple yet effective CNN-based pipeline that does not require any region analysis. We demonstrate the generalization of our approach across different retrieval domains, resulting in a scalable retrieval system that does not require any additional feature finetuning. 

\section{Bags of Local Convolutional Features}
\label{sec:BLCF}
Our pipeline extends the work proposed in~\cite{mohedano2016bags}, that consists of using the traditional bag of visual words encoding on local pre-trained convolutional features. The bags of local convolutional features (BLCF) has the advantage of generating high-dimensional and sparse representations. The high-dimensionality makes the aggregated local information more likely to be linearly separable, while  relatively few non-zero elements means that they are efficient in terms of storage and computation, as shown in Table~\ref{tab:memtime}. 

\begin{table}[t!]
\center
\caption{Average number of visual words per image in 25k BoW representation on different benchmarks. We also report the query time (seconds).\vspace{0.1cm}}
\label{tab:memtime}
\begin{tabular}{@{}p{3.7cm}cccc@{}}
\toprule
               & INSTRE & Oxford & Paris  \\ \midrule
words/image    & 289 & 174  & 163     \\
query/time  & 0.019 & 0.002 & 0.003         \\ \bottomrule
\end{tabular}
\end{table}

\begin{figure}[b!]
\begin{center}
\includegraphics[width=.48\textwidth]{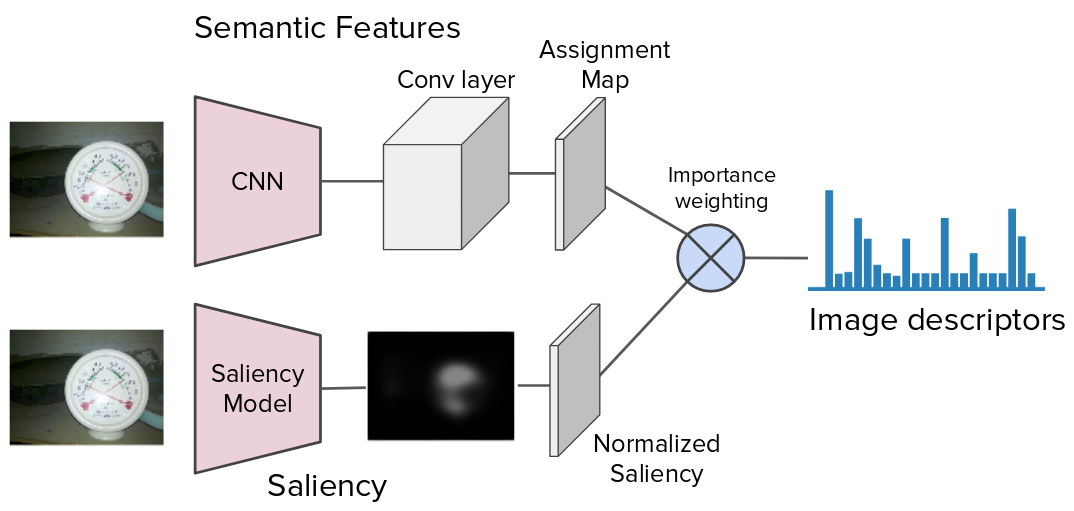}
\caption{The Bag of Local Convolutional Framework (BLCF) pipeline with saliency weighting.}
\label{fig:BLCF_saliency}
\end{center}
\end{figure}

The volume of activations from a convolutional layer $\mathcal{X}$ in $\mathbb{R}^{M \times N \times D}$, is re-interpreted as a set of $M \times N$ local descriptors of $D$ dimensions. A visual vocabulary is learned using $k$-means on the local CNN features. With this procedure each image is represented with an \textit{assignment map}, which consists of a compact representation of a 2D array of size $H \times W$ relating each pixel of the original image with its visual work with precision of ($W/N$, $H/M$); being $W$ and $H$ the the with and height of the original images. This is represented in the first stream of Figure~\ref{fig:BLCF_saliency}, which involves the semantic representation of the image.

One of the main advantages of this representation is to preserve the spatial layout of the image, so it is possible to apply a spatial weighting scheme prior to constructing the final BoW representation. In particular, a saliency weighting scheme $\alpha(i,j)$ with resolution $W \times H$ is down-sampled to match the spatial resolution of the assignment maps $M \times N$, and normalized to have values between 0 and 1, illustrated in the second stream in Figure~\ref{fig:BLCF_saliency}. Then, the final representation consists in an histogram $\mathbf{f}_{\text{BOW}}=(h_{1}, ... h_{K})$, where each component is the sum of the spatial weight $w_{ik}$ assigned to a particular visual word $k$, $h_{k}=\sum_{i=1}^N{w_{ik}}$. The final vector is $L^2$-normalized and cosine similarity is used to rank the images.

We use descriptors from \textit{conv5\_1} from a pre-trained VGG16~\cite{vgg} without fully connected layers. The VGG16 network was chosen following Jimenez et al. ~\cite{Jimenez_2017_BMVC}, where it was found that the features from this network outperform those from ResNet50 using the CAM approach. We confirmed this result in our own experiments, finding a similar result for BCLF encoding when using the local search scheme: there was a consistent reduction ($>$ 10 points) in mAP when compared with VGG16 regardless of the weighting scheme used. Images are resized to have its maximum size of 340 pixels ($1/3$ of the original resolution) before performing mean subtraction prior to being forwarded through the network. This resolution is also used to extract the different saliency maps. Local features are post-processed with $L^2$-normalization, followed by a PCA whitening (512D), and $L^2$-normalized again prior to being clustered. Clustering is performed using approximate $k$-means with $k=25,000$ words. The visual vocabulary and PCA models are fit on the target dataset. The obtained saliency maps are down-sampled by computing the maximum across $16\times 16$ non-overlapped blocks.


For the query images, we interpolate the volume of activations to $(2H,2W)$, as this interpolation of the feature maps has been shown to improve performance~\cite{mohedano2016bags}. However, we only apply this strategy to the query images, and not to the entire dataset, so as not to increase dataset memory consumption. The bounding box of the queries is mapped to the assignment maps, and only the information relevant to the instance is used to create $\mathbf{f}_{\text{BOW}}$. This procedure is similar to the one used in~\cite{Radenović2016}, where a bounding box is mapped to the convolutional feature maps instead of cropping the original image, which was shown to provide better results.

\section{Saliency models}
\label{sec:saliency}
\begin{figure*}[t]
\begin{center}
\includegraphics[width=1.0\textwidth]{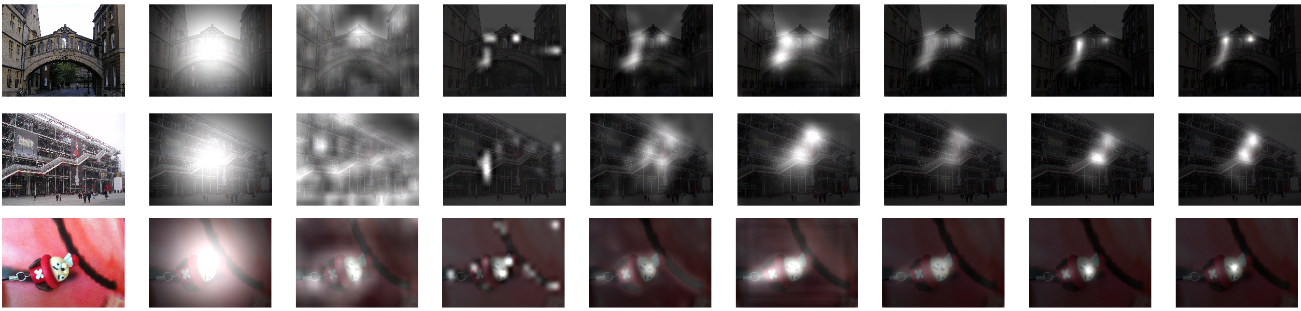}
\caption{Three sample query images from Oxford (top row), Paris (middle row), and INSTRE (bottom row) with their associated spatial weighting schemes: From left to right weighting schemes based on Gaussian center prior, $L^2$-norm of convolutional features, IttiKoch model, BMS, SalNet, SalGAN, SAM-VGG, and SAM-ResNet.}
\label{fig:saliency_models}
\end{center}
\end{figure*}

One of the main features of the human vision system is its capacity to actively focus on the most salient regions and movements. Visual attention models detect regions that stand out from their surroundings, producing saliency maps that highlight the most prominent regions within an image. Saliency has been shown beneficial in content-based image retrieval when using traditional the Bag of Words model and SIFT features~\cite{Wu2017}. Some works have investigated the usage of saliency as a mechanism to reduce the number of local features and reduce the computational complexity of SIFT-based BoW frameworks~\cite{awad2012saliency,liang2010salient,nakamoto2011combination}. Other works, rather than completely discard the background information, have used saliency to weight the contribution of the foreground and the background simultaneously~\cite{de2012spatial,wu2017visual}. However, the usage of saliency models in the task of image retrieval has thus far been restricted to handcrafted CBIR features and handcrafted saliency models~\cite{itti1998model}.

With the emergence of challenges such as the MIT saliency benchmark~\cite{MIT300} and the Large-Scale Scene Understanding Challenge~\cite{LSUNchallenge} (LSUN), and the appearance of large-scale public annotated datasets such as iSUN~\cite{iSUN}, SALICON~\cite{SALICON}, MIT300~\cite{MIT300}, or CAT2000~\cite{CAT2000}, data-driven approaches based on CNN models trained end-to-end have become dominant, generating more accurate models each year~\cite{Deepgaze,SALICON,Deepfix,SalGAN,SalNet}. Having access to large datasets  has not only allowed the development of deep learning based saliency models, but has also enabled a better comparison of the performance between models.

Our goals are to first evaluate whether saliency is useful in a CNN-based retrieval system, and second, to examine how the performance of different saliency models on saliency benchmarks relates to performance of  instance search. To this end, we select two handcrafted models and four deep learning based models. The handcrafted models selected are the classic approach proposed by Itti and Koch~\cite{itti1998model}, and the Boolean Map based saliency (BMS)~\cite{zhang2013saliency}, which represents the best performing non-deep learning algorithm on the MIT300 benchmark. The four deep learning based models differ in CNN architecture, the loss functions used in training, and the training strategy~\cite{SalNet,SalGAN,cornia2016predicting}, providing a reasonable cross section of the current state-of-the-art. Table~\ref{tab:MITbench} summarizes the saliency prediction performance of the selected models, where we report the the AUC-Judd value (which is variant of the Area Under the Curve), the Pearson's linear correlation coefficient (CC), the Normalized Scanpath Saliency (NSS), and the Kullback-Leibler divergence (KL) to evaluate how accurate the saliency prediction is when comparing with the eye-tracking ground truth in the MIT300 benchmark.

\begin{table}[b]
\centering
\caption{Performance of CNN saliency models on the MIT300 saliency benchmark~\cite{MIT300}.\vspace{0.1cm}}
\label{tab:MITbench}
\begin{tabular}{@{}p{3.9cm}llll@{}}
\toprule
Saliency model & \begin{tabular}[c]{@{}l@{}}AUC\\ Judd\end{tabular} & CC   & NSS  & KL   \\ \midrule
Human          & 0.92                                               & 1.00    & 3.29 & 0.00    \\
SAM-ResNet~\cite{cornia2016predicting}     & 0.87                                               & 0.78 & 2.34 & 1.27 \\
SAM-VGG~\cite{cornia2016predicting}        & 0.87                                               & 0.77 & 2.30 & 1.13 \\
SalGAN~\cite{SalGAN}         & 0.86                                               & 0.73 & 2.04 & 1.07 \\
SalNet~\cite{SalNet}         & 0.83                                               & 0.58 & 1.51 & 0.81 \\
BMS~\cite{zhang2013saliency}         &        0.83                                        & 0.55 & 1.41 & 0.81 \\ 
IttiKoch~\cite{walther2006modeling}         & 0.60                                               & 0.14 & 0.43 & 2.30 \\ \bottomrule
\end{tabular}
\end{table}

The \textbf{Itti-Koch model}~\cite{walther2006modeling} model is inspired by how the primary visual cortex processes visual information. It explicitly models local variations on three early visual features to generate the final saliency maps. A salient region represents a region presenting a clear contrast with its surroundings in terms of color, intensity, and/or orientation. The proposed algorithm analyzes the image at eight different scales and computes center surround difference between fine and coarse levels. Three ``conspicuity'' maps, one for each visual feature, are obtained by normalizing and performing across-scale addition. The final saliency map is generated by a linear combination of the three normalized conspicuity maps. The final step of the model includes a winner-take-all (WTA) network that simulates a scan path over the image in order of decreasing saliency of attended locations. We use the MATLAB implementation from Walther and Koch\footnote{http://www.saliencytoolbox.net/}, which extends the work of Itti-Koch~\cite{itti1998model} to generate saliency regions with the extent of the objects contained in the image.

\textbf{Boolean Map based Saliency}~\cite{zhang2013saliency} generates a set of Boolean maps by randomly thresholding the different color channels of an image. Then, an attention map for each Boolean map is computed by assigning 1 to the union of surrounded regions, and 0 to the rest of the map. The attention maps are normalized (with a dilatation operation and a $L^2$-normalization) and linearly combined into a full-resolution mean attention map. A Gaussian blur is applied on the obtained map generating the final saliency prediction. Despite its simplicity, this method achieves the best performance among other hand-crafted approaches in the saliency MIT300 benchmark.

\textbf{Deep Salnet}~\cite{SalNet} is a end-to-end deep learning model trained to predict human eye fixations. It uses a fully convolutional network consisting of ten layers: one input layer, eight convolution layers, and one deconvolution layer~\cite{deconvolution}. The ReLU is used as an activation function and pooling layers follow the first two convolutional layers, effectively reducing the width and height of the feature maps in the intermediate layers by a factor of four. The final deconvolution generates the final prediction, up-sampling the resolution to the original input size. A transfer learning strategy is applied on this network by re-using and adapting the weights from the first three convolutional layers from a pre-trained VGG architecture~\cite{vgg}. This acts as a regularizer and improves the final network result. The loss function used during training is the mean squared error computed pixel-wise between the saliency prediction and the ground truth map.

\begin{figure*}[t]
\begin{center}
\includegraphics[width=1.0\textwidth]{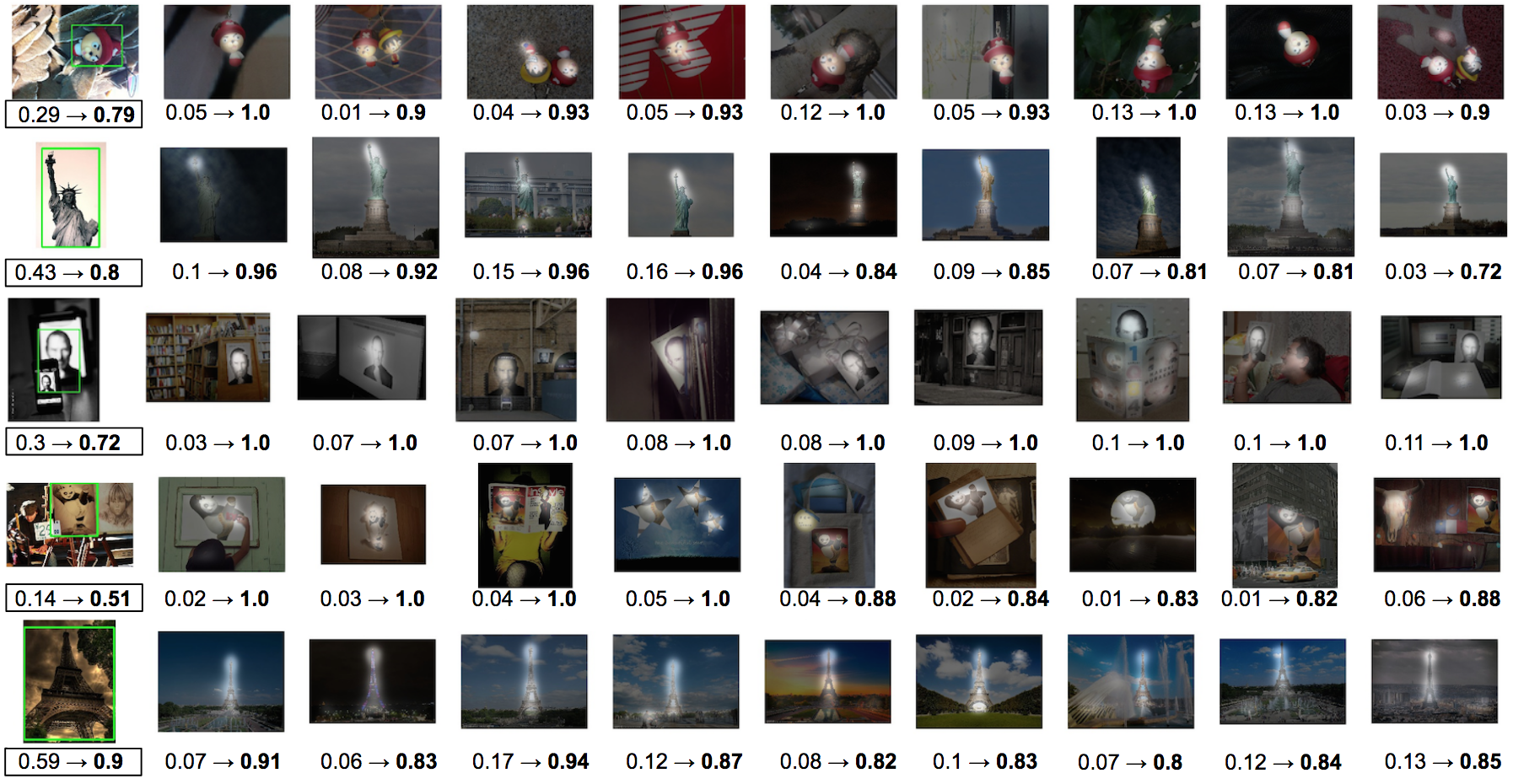}
\caption{Effect on performance for five examples from the INSTRE dataset after applying saliency weighting (SalGAN). The first column depicts the query with its associated average precision (AP). On the left, AP when performing unweighted BLCF, and on the right when performing saliency weighted BLCF. Retrieved images are ranked in decreasing order of ranking difference between unweighted BLCF and saliency weighted BLCF. For each image, precision at the position where each image is retrieved is reported for unweighted BLCF (left) and saliency weighted BLCF (right).}
\label{fig:saliency_models}
\end{center}
\end{figure*}

The \textbf{SalGAN}~\cite{SalGAN} model introduces the use of adversarial training~\cite{goodfellow2014generative} for the task of visual saliency prediction. The model is the \textit{generator} module in the adversarial setup. It follows the VGG-16~\cite{vgg} architecture without the fully connected layers, and it is initially trained using binary cross entropy loss to predict eye fixations, similarly to~\cite{SalNet}. The \textit{decoder} architecture is structured in the same way as the encoder, but in reverse order and with pooling layers replaced by upsampling layers. The weights of the decoder are randomly initialized. The final loss used for the adversarial training is a combination of the a content loss (binary-cross entropy of predicted saliency maps and the saliency ground truth) and an adversarial loss quantifying the error of the discriminator.

The \textbf{Saliency Attentive Model}~\cite{cornia2016predicting} is a CNN model whose distinctive feature is to include an attentive mechanism based on Long-Short Term Memory (LSTM) to iteratively refine predicted saliency maps. The authors also explicitly model the tendency of looking towards the center of images~\cite{tatler2007central} by including a set of trainable Gaussian priors on their model. The base CNN model uses dilated convolutions~\cite{yu2015multi} to better preserve the spatial information across layers, and its weights are initialized on ImageNet. Parameters of the attention mechanism and center-priors are fit for saliency prediction using a loss function that consists of a combination of three popular salience estimation measurements (NSS, CC, and KL). Two models are proposed, one based on a VGG16-like architecture~\cite{vgg} (SAM-VGG), and a second based on ResNet-50~\cite{He2016DeepRL} (SAM-ResNet).

Figure~\ref{fig:saliency_models} shows the different spatial weights evaluated in this work. Additionally, we evaluate a fixed spatial weighting approach (Gaussian center-prior~\cite{BabenkoL15}) and a weighting scheme derived from the activations of the convolutional layer (by measuring the $L^2$-norm of each local CNN representation similar to~\cite{CroW}).

\section{Experiments}
\label{sec:experiments}

We experiment with three different datasets: the challenging INSTRE dataset for object retrieval ~\cite{Instre}, and two well-known landmark-related benchmarks, the Oxford~\cite{Oxford5k} and Paris~\cite{paris6k} datasets. The former dataset was specifically designed to evaluate instance retrieval approaches. It consists in 23,070 manually annotated images, including 200 different instances from diverse domains, including logos, 3D objects, buildings, and sculptures. Performance is evaluated using mean average precision (mAP) following the same standard protocols as in Oxford and Paris benchmarks, and evaluating mAP over 1,200 images as described in~\cite{difussion}.

\subsection{Weighting schemes}
\label{sub:wschemes}
\begin{table}[b!]
\caption{Performance (mAP) of different spatial weighting schemes using the BLCF approach.\vspace{0.1cm}}
\centering
\begin{tabular}{@{}p{3.9cm}lll@{}}
\toprule
Weighting           & INSTRE         & Oxford         & Paris          \\ \midrule
None       & 0.636          & 0.722          & 0.798          \\
Gaussian   & 0.656          & 0.728          & 0.809          \\
$L^2$-norm & 0.674          & 0.740          & \textbf{0.817} \\
Itti-Koch~\cite{itti1998model}  & 0.633          & 0.693          & 0.785          \\
BMS~\cite{zhang2013saliency}        & 0.688          & 0.729          & 0.806          \\
SalNet~\cite{SalNet}     & 0.688          & 0.746          & 0.814          \\
SalGAN~\cite{SalGAN}     & \textbf{0.698} & \textbf{0.746} & 0.812          \\
SAM-VGG~\cite{cornia2016predicting}    & 0.688          & 0.686          & 0.785          \\
SAM-ResNet~\cite{cornia2016predicting} & 0.688          & 0.673          & 0.780          \\ \bottomrule
\end{tabular}
\label{tab:resultsBLCF}
\end{table}

In this section we evaluate different spatial weighting schemes with the BLCF framework. We find that saliency weighting schemes (BMS, SalNet, and SalGAN) improve retrieval performance across the evaluated datasets with respect to the unweighted assignment maps. Figure~\ref{fig:saliency_models} shows some of the cases where saliency (SalGAN model) is most beneficial, allowing the efficient localisation of the target instances in most of the cases, despite of the high variability of the backgrounds and relative positions within the images in the INSTRE dataset.

Table~\ref{tab:resultsBLCF} contains the performance of different weighting schemes on the BLCF approach. The simple Gaussian weighting achieves a boost in performance with respect to the baseline, which indicates a natural tendency of the instances to appear in the center of the images in all the datasets. The saliency measure derived from the convolutional features ($L^2$-norm weighting) allows to roughly localize the most relevant parts of the image, which represents a boost in performance with respect the Gaussian center prior weighting. However, the $L^2$ weighting  appears in general to be much more noisy than the human attention-based saliency models, as depicted in Figure~\ref{fig:weighting_schemes} and Figure~\ref{fig:saliency_models}, which leads to lower mAP when compared with the BMS, SalNet, and SalGAN models.

Saliency models achieve the best performing results in the INSTRE dataset, with the exception of the Itti-Koch model, which decreases performance with respect the baseline in all datasets. This result is consistent with the quality of the saliency prediction achieved in the MIT300 benchmark, where it is rated in the second lowest rank of the evaluated models~\footnote{\url{http://saliency.mit.edu/results_mit300.html}}. The more accurate saliency models (BMS, SalNet, SalGAN, SAM-VGG, and SAM-ResNet), however, achieve almost equivalent performance on the INSTRE dataset. The relatively small size of the instances within this dataset and the high variance in their relative position within the images makes saliency a very efficient spatial weighting strategy for instance search, with no substantial difference between saliency models.

\begin{figure}[t!]
\begin{center}
\includegraphics[width=.48\textwidth]{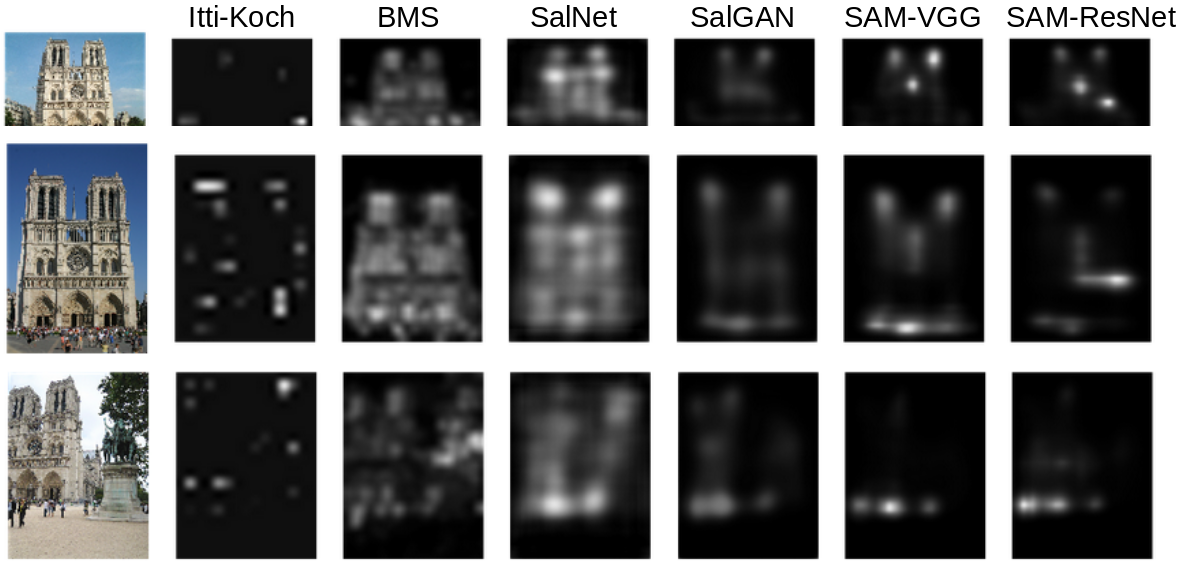}
\caption{\small Sample of saliency predictions for three examples of Notre Dame cathedral in the Paris dataset. Average precision (AP) for this query is 0.868, which is improved by BMS and SalNet models (achieving 0.880 and 0.874 AP respectively). More accurate saliency models decrease performance with respect the baseline, achieving an AP of 0.862 in the case of SalGAN, 0.857 SAM-VGG, 0.856 SAM-ResNet and 0.853 Itti-Koch models.}
\label{fig:Notredameexample}
\end{center}
\end{figure}

This contrasts with results for the Oxford and Paris datasets, where a coarser saliency prediction (i.e the one provided by the SalNet model) achieves better results than the one obtained with the more accurate models of SAM-VGG and SAM-ResNet. Figure~\ref{fig:Notredameexample} illustrates this effect on three different instances of the Notre Dame cathedral from the Paris dataset. The most accurate saliency models (i.e. SAM-VGG and SAM-ResNet) detect saliency regions ``within'' the building, instead of detecting the full building as relevant. Also, SAM-VGG, SAM-ResNet, and SalGAN appear to be more sensitive to detecting people, omitting any other salient region of the image and thus decreasing the overall retrieval performance.

\subsection{Aggregation methods}
We also tested saliency weighting in combination with classic sum pooling over the spatial dimension to compare with the BCLF weighting. Here we report the best performing saliency models (SalNet and SalGAN) compared with a no weighting baseline using the VGG-16 network \textit{pool5} layer at full image resolution. Although combining sum pooling with saliency weighting did give a performance improvement on the INSTRE dataset (mAP 0.527 for SalGAN vs 0.405 for no pooling), saliency weighting combined with BLCF gives substantially better mAP (see Table~\ref{tab:aggregation}). Furthermore, sum pooling with saliency weighting gave little or no improvement on the Oxford and Paris datasets (mAP Oxford: 0.681 for SalNet vs 0.686 for no weighting, and Paris: 0.766 for SalNet vs 0.765 for no weighting). This result is perhaps unsurprising. Weighting the feature assignments prior to pooling in the BCLF framework can be interpreted as a simple importance weighting on each discrete feature. However, the interpretation for sum pooling is less clear, since the feature vectors at each spatial location usually have multiple activations of varying magnitude, and weighting alters the potentially semantically meaningful magnitudes of the activations. The BCLF-saliency approach also has the advantage of having two to three times fewer non-zero elements in the representation, giving substantially improved query times.

\begin{table}[t!]
\caption{Performance (mAP) of different spatial weighting schemes using the sum pooling aggregation on (SUM) and the BLCF approach.\vspace{0.1cm}}
\centering
\begin{adjustbox}{width=0.48\textwidth}
\begin{threeparttable}
\begin{tabular}{@{}lllllll@{}}
\toprule
Method            & \multicolumn{2}{c}{Instre} & \multicolumn{2}{c}{Oxford} & \multicolumn{2}{c}{Paris} \\ \midrule
Weighting & SUM          & BLCF        & SUM          & BLCF        & SUM         & BLCF        \\
None              & 0.405        & 0.636       & 0.686        & 0.722       & 0.765       & 0.798       \\
SalNet            & 0.519        & 0.688       & 0.681        & 0.746       & 0.766       & 0.814       \\
SalGAN            & 0.527        & 0.698       & 0.612        & 0.746       & 0.749       & 0.812       \\
\end{tabular}
\end{threeparttable}
\end{adjustbox}
\label{tab:aggregation}
\end{table}

\subsection{Comparison with the state-of-the-art}

Our approach compares favorably with other methods exploiting pre-trained convolutional features, as shown in Table~\ref{tab:SOA}. In particular, we achieve state-of-the-art performance using pre-trained features in Oxford and INSTRE datasets, when combining BLCF with saliency prediction from SalGAN. CAMs~\cite{Jimenez_2017_BMVC}, and R-MAC~\cite{rmac-ToliasSJ15} slightly outperform our approach in the Paris dataset, where they achieve $0.855$, and $0.835$ mAP respectively, whereas our method achieves $0.812$ mAP. It is worth noting, however, that our system is much less complex (we do not perform region analysis), and more scalable, since the sparse representation only needs a maximum of $336$ non-zero elements out of the 25K dimensions.

Fine-tuned CNN models~\cite{Radenović2016,gordo2016deep} significantly outperform our approach in Oxford and Paris. However, saliency weighted BLCF achieves state-of-the art performance in the INSTRE dataset with $0.698$ mAP, outperforming the fined-tuned models. This is probably a consequence of restricting the training domain to landmarks images in~\cite{Radenović2016,gordo2016deep}, since R-MAC with pre-trained features~\cite{rmac-ToliasSJ15} ($0.523$ mAP) already achieves a higher performance in INSTRE than the fine-tuned R-MAC~\cite{gordo2016deep} ($0.477$ mAP). This provides evidence that similarity learning is a strategy to greatly improve performance on CNN representations where the target instance domain is restricted. In more generic scenarios, such as the INSTRE dataset, pre-trained ImageNet models achieve more general image representations.

One common technique to improve retrieval results is the average query expansion (AQE)~\cite{chum2007total}. Given a list of retrieved images, a new query is generated by sum aggregating the image representations of the top $N$ images for a particular query. We select the top 10 images, and $L^2$-normalization on the new query representation. Table~\ref{tab:SOA_QE} With this simple post processing we substantially improve the results in the INSTRE dataset (mAP increased from 0.698 to 0.757), achieving the best performance in comparison with other methods using the same post-processing strategy.

\begin{table}[t!]
\centering
\caption{Performance comparison with the state-of-the-art.\vspace{0.1cm}}
\label{tab:SOA}
\begin{adjustbox}{width=0.48\textwidth}
\begin{threeparttable}
\begin{tabular}{lccccc}
\hline
Method       & \multicolumn{1}{l}{Off-the-shelf} & dim             & INSTRE         & Oxford         & Paris 
        \\ \hline
CroW~\cite{CroW}         & yes                               & 512             & 0.416          & 0.698          & 0.797          \\
CAM~\cite{Jimenez_2017_BMVC}\tnote{*}        & yes                               & 512             &       0.325         & 0.736          & 0.855          \\
R-MAC~\cite{rmac-ToliasSJ15}        & yes                               & 512             & 0.523          & 0.691          & 0.835          \\
R-MAC~\cite{Radenović2016}\tnote{\dag}        & No                                & 512             & 0.477          & 0.777          & 0.841          \\
R-MAC-ResNet~\cite{gordo2016deep}\tnote{\dag} & No                                & 2048            & 0.626          & \textbf{0.839} & \textbf{0.938} \\
(our) BLCF  & yes                               & 336 & 0.636 &    0.722      &    0.798       \\
(our) BLCF-Gaussian  & yes                               & 336 & 0.656 & 0.728          &    0.809       \\
(our) BLCF-SalGAN  & yes                               & 336 & \textbf{0.698} & 0.746          & 0.812          \\
\end{tabular}
\begin{tablenotes}
	\centering
    \footnotesize
     \item Results marked with (*) are provided by the authors. Those marked with (\dag) are reported in Iscen et al.~\cite{difussion}. Otherwise they are based on our own implementation.
    \end{tablenotes}
\end{threeparttable}
\end{adjustbox}
\end{table}

\begin{table}[t!]
\centering
\caption{Performance comparison with the state-of-the-art with average query expansion.\vspace{0.1cm}}
\label{tab:SOA_QE}
\begin{adjustbox}{width=0.48\textwidth}
\begin{threeparttable}
\begin{tabular}{@{}lccccc@{}}
\toprule
Method       & \multicolumn{1}{l}{Off-the-shelf} & dim             & INSTRE         & Oxford         & Paris          \\ \midrule
CroW~\cite{CroW}         & yes                               & 512             & 0.613          & 0.741          & 0.855          \\
CAM~\cite{Jimenez_2017_BMVC}\tnote{*}          & yes                               & 512             &                & 0.760          & 0.873          \\
R-MAC~\cite{rmac-ToliasSJ15}        & yes                               & 512             & 0.706          & 0.770          & 0.884          \\
R-MAC~\cite{Radenović2016}\tnote{\dag}        & No                                & 512             & 0.573          & 0.854          & 0.884          \\
R-MAC-ResNet~\cite{gordo2016deep}\tnote{\dag} & No                                & 2048            & 0.705          & \textbf{0.896} & \textbf{0.953} \\
(ours) BLCF  & yes                               & 336 & 0.679 & 0.751          & 0.788          \\
(ours) BLCF-Gaussian  & yes                               & 336 & 0.731 & 0.778          & 0.838          \\
(ours) BLCF-SalGAN  & yes                               & 336 & \textbf{0.757} & 0.778          & 0.830          \\
\end{tabular}
\begin{tablenotes}
	\centering
    \footnotesize
     \item Results marked with (*) are provided by the original publications. Those marked with (\dag) are reported in Iscen et al.~\cite{difussion}. Otherwise they are based on our own implementation.
    \end{tablenotes}
\end{threeparttable}
\end{adjustbox}
\end{table}

\section{Conclusions and future work}
\label{sec:conclusions}

We have proposed a generic method for instance search that relies on BoW encoding of local convolutional features and attention models. We have demonstrated that saliency models are useful for the instance search task using a recently proposed CNN-based retrieval framework. Results indicate that different state-of-the-art saliency prediction models are equally beneficial for this task in the challenging INSTRE dataset. In landmark related datasets such as Oxford an Paris, however, a coarse saliency is more beneficial than highly accurate saliency models such as SAM-VGG or SAM-ResNet.

Better query expansion strategies can be applied to further improve results, such as the diffusion strategy proposed in~\cite{difussion}, where global diffusion using $2,048$ dimensional descriptors from fine tuned R-MAC~\cite{gordo2016deep} achieves $0.805$ mAP in INSTRE, and a regional diffusion $0.896$ mAP. This strategy can be also applied to the proposed saliency weighted BLCF representation, potentially increasing retrieval performance of the proposed saliency weighted BLCF representations.
\section*{Acknowledgments}
\noindent
This publication has emanated from research conducted with the financial support of Science Foundation Ireland (SFI) under grant number SFI/12/RC/2289 and SFI/15/SIRG/3283. This research was supported by contract SGR1421 by the Catalan AGAUR office.
The work has been developed in the framework of project TEC2016-75976-R, funded by the Spanish Ministerio de Economia y Competitividad and the European Regional Development Fund (ERDF). The authors also thank NVIDIA for generous hardware donations.

{\small
\bibliographystyle{ieee}
\bibliography{egbib}

\begin{thebibliography}{10}\itemsep=-1pt

\bibitem{awad2012saliency}
D.~Awad, V.~Courboulay, and A.~Revel.
\newblock Saliency filtering of sift detectors: Application to cbir.
\newblock In {\em International Conference on Advanced Concepts for Intelligent
  Vision Systems}, pages 290--300. Springer, 2012.

\bibitem{Awad2017}
G.~Awad, W.~Kraaij, P.~Over, and S.~Satoh.
\newblock Instance search retrospective with focus on {TRECVID}.
\newblock {\em International Journal of Multimedia Information Retrieval},
  6(1):1--29, Mar 2017.

\bibitem{neuralcodes}
A.~Babenko, A.~Slesarev, A.~Chigorin, and V.~Lempitsky.
\newblock Neural codes for image retrieval.
\newblock In {\em Computer Vision--ECCV 2014}, pages 584--599. 2014.

\bibitem{CAT2000}
A.~Borji and L.~Itti.
\newblock Cat2000: A large scale fixation dataset for boosting saliency
  research.
\newblock {\em arXiv preprint arXiv:1505.03581}, 2015.

\bibitem{MIT300}
Z.~Bylinskii, T.~Judd, A.~Borji, L.~Itti, F.~Durand, A.~Oliva, and A.~Torralba.
\newblock Mit saliency benchmark.

\bibitem{CaoLWHSS16}
J.~Cao, L.~Liu, P.~Wang, Z.~Huang, C.~Shen, and H.~T. Shen.
\newblock Where to focus: Query adaptive matching for instance retrieval using
  convolutional feature maps.
\newblock {\em arXiv preprint arXiv:1606.06811}, 2016.

\bibitem{chum2007total}
O.~Chum, J.~Philbin, J.~Sivic, M.~Isard, and A.~Zisserman.
\newblock Total recall: Automatic query expansion with a generative feature
  model for object retrieval.
\newblock In {\em International Conference on Computer Vision (ICCV)}, pages
  1--8, 2007.

\bibitem{cornia2016predicting}
M.~Cornia, L.~Baraldi, G.~Serra, and R.~Cucchiara.
\newblock Predicting human eye fixations via an lstm-based saliency attentive
  model.
\newblock {\em arXiv preprint arXiv:1611.09571}, 2016.

\bibitem{de2012spatial}
R.~de~Carvalho~Soares, I.~R. da~Silva, and D.~Guliato.
\newblock Spatial locality weighting of features using saliency map with a
  bag-of-visual-words approach.
\newblock In {\em Tools with Artificial Intelligence (ICTAI), 2012 IEEE 24th
  International Conference on}, volume~1, pages 1070--1075. IEEE, 2012.

\bibitem{goodfellow2014generative}
I.~Goodfellow, J.~Pouget-Abadie, M.~Mirza, B.~Xu, D.~Warde-Farley, S.~Ozair,
  A.~Courville, and Y.~Bengio.
\newblock Generative adversarial nets.
\newblock In {\em Advances in neural information processing systems}, pages
  2672--2680, 2014.

\bibitem{gordo2016deep}
A.~Gordo, J.~Almaz{\'{a}}n, J.~Revaud, and D.~Larlus.
\newblock End-to-end learning of deep visual representations for image
  retrieval.
\newblock {\em International Journal of Computer Vision}, 124(2):237--254,
  2017.

\bibitem{He2016DeepRL}
K.~He, X.~Zhang, S.~Ren, and J.~Sun.
\newblock Deep residual learning for image recognition.
\newblock {\em 2016 IEEE Conference on Computer Vision and Pattern Recognition
  (CVPR)}, pages 770--778, 2016.

\bibitem{difussion}
A.~Iscen, G.~Tolias, Y.~Avrithis, T.~Furon, and O.~Chum.
\newblock Efficient diffusion on region manifolds: Recovering small objects
  with compact cnn representations.
\newblock In {\em 2017 IEEE Conference on Computer Vision and Pattern
  Recognition (CVPR)}, 2017.

\bibitem{itti1998model}
L.~Itti, C.~Koch, and E.~Niebur.
\newblock A model of saliency-based visual attention for rapid scene analysis.
\newblock {\em IEEE Transactions on pattern analysis and machine intelligence},
  20(11):1254--1259, 1998.

\bibitem{vlad}
H.~J{\'e}gou, M.~Douze, C.~Schmid, and P.~P{\'e}rez.
\newblock Aggregating local descriptors into a compact image representation.
\newblock In {\em Computer Vision and Pattern Recognition (CVPR)}, pages
  3304--3311, 2010.

\bibitem{SALICON}
M.~Jiang, S.~Huang, J.~Duan, and Q.~Zhao.
\newblock Salicon: Saliency in context.
\newblock In {\em 2015 IEEE Conference on Computer Vision and Pattern
  Recognition (CVPR)}, pages 1072--1080, June 2015.

\bibitem{Jimenez_2017_BMVC}
A.~Jimenez, J.~M. Alvarez, and X.~Giro-i Nieto.
\newblock Class-weighted convolutional features for visual instance search.
\newblock In {\em 28th British Machine Vision Conference (BMVC)}, September
  2017.

\bibitem{tembedding}
H.~Jégou and A.~Zisserman.
\newblock Triangulation embedding and democratic aggregation for image search.
\newblock In {\em 2014 IEEE Conference on Computer Vision and Pattern
  Recognition}, pages 3310--3317, June 2014.

\bibitem{CroW}
Y.~Kalantidis, C.~Mellina, and S.~Osindero.
\newblock Cross-dimensional weighting for aggregated deep convolutional
  features.
\newblock {\em CoRR}, abs/1512.04065, 2015.

\bibitem{Deepfix}
S.~S. Kruthiventi, K.~Ayush, and R.~V. Babu.
\newblock Deepfix: A fully convolutional neural network for predicting human
  eye fixations.
\newblock {\em IEEE Transactions on Image Processing}, 2017.

\bibitem{Deepgaze}
M.~K{\"{u}}mmerer, L.~Theis, and M.~Bethge.
\newblock Deep gaze {I:} boosting saliency prediction with feature maps trained
  on imagenet.
\newblock {\em CoRR}, abs/1411.1045, 2014.

\bibitem{QAwrmac17}
Z.~Laskar and J.~Kannala.
\newblock Context aware query image representation for particular object
  retrieval.
\newblock In {\em Scandinavian Conference on Image Analysis}, pages 88--99.
  Springer, 2017.

\bibitem{liang2010salient}
Z.~Liang, H.~Fu, Z.~Chi, and D.~Feng.
\newblock Salient-sift for image retrieval.
\newblock In {\em International Conference on Advanced Concepts for Intelligent
  Vision Systems}, pages 62--71. Springer, 2010.

\bibitem{mohedano2016bags}
E.~Mohedano, K.~McGuinness, N.~E. O'Connor, A.~Salvador, F.~Marqu{\'e}s, and
  X.~Gir{\'o}-i Nieto.
\newblock Bags of local convolutional features for scalable instance search.
\newblock In {\em Proceedings of the 2016 ACM on International Conference on
  Multimedia Retrieval}, pages 327--331. ACM, 2016.

\bibitem{nakamoto2011combination}
S.~Nakamoto and T.~Toriu.
\newblock Combination way of local properties classifiers and saliency in
  bag-of-keypoints approach for generic object recognition.
\newblock {\em International Journal of Computer Science and Network Security},
  11(1):35--42, 2011.

\bibitem{SalGAN}
J.~Pan, C.~Canton{-}Ferrer, K.~McGuinness, N.~E. O'Connor, J.~Torres,
  E.~Sayrol, and X.~{Gir{\'{o}} i Nieto}.
\newblock Salgan: Visual saliency prediction with generative adversarial
  networks.
\newblock {\em CoRR}, abs/1701.01081, 2017.

\bibitem{SalNet}
J.~Pan, E.~Sayrol, X.~Giro-i Nieto, K.~McGuinness, and N.~E. O'Connor.
\newblock Shallow and deep convolutional networks for saliency prediction.
\newblock In {\em Proceedings of the IEEE Conference on Computer Vision and
  Pattern Recognition}, pages 598--606, 2016.

\bibitem{Oxford5k}
J.~Philbin, O.~Chum, M.~Isard, J.~Sivic, and A.~Zisserman.
\newblock Object retrieval with large vocabularies and fast spatial matching.
\newblock In {\em Proceedings of the IEEE Conference on Computer Vision and
  Pattern Recognition}, 2007.

\bibitem{paris6k}
J.~Philbin, O.~Chum, M.~Isard, J.~Sivic, and A.~Zisserman.
\newblock Lost in quantization: {I}mproving particular object retrieval in
  large scale image databases.
\newblock In {\em Proceedings of the IEEE Conference on Computer Vision and
  Pattern Recognition}, 2008.

\bibitem{Radenović2016}
F.~Radenovi{\'c}, G.~Tolias, and O.~Chum.
\newblock {CNN} image retrieval learns from {BoW}: Unsupervised fine-tuning
  with hard examples.
\newblock In {\em European Conference on Computer Vision}, pages 3--20.
  Springer International Publishing, 2016.

\bibitem{cnnofftheshelf}
A.~S. Razavian, H.~Azizpour, J.~Sullivan, and S.~Carlsson.
\newblock {CNN} features off-the-shelf: an astounding baseline for recognition.
\newblock In {\em Computer Vision and Pattern Recognition Workshops (CVPRW)},
  2014.

\bibitem{visualInstanceRazavianSMC14}
A.~S. Razavian, J.~Sullivan, S.~Carlsson, and A.~Maki.
\newblock Visual instance retrieval with deep convolutional networks.
\newblock {\em ITE Transactions on Media Technology and Applications},
  4(3):251--258, 2016.

\bibitem{FastERGirshick15}
S.~Ren, K.~He, R.~Girshick, and J.~Sun.
\newblock Faster r-cnn: Towards real-time object detection with region proposal
  networks.
\newblock {\em IEEE Transactions on Pattern Analysis and Machine Intelligence},
  39(6):1137--1149, June 2017.

\bibitem{simeoni:unsupervisedObject}
O.~Sim{\'e}oni, A.~Iscen, G.~Tolias, Y.~Avrithis, and O.~Chum.
\newblock Unsupervised deep object discovery for instance recognition.
\newblock {\em arXiv preprint arXiv:1709.04725}, 2017.

\bibitem{vgg}
K.~Simonyan and A.~Zisserman.
\newblock Very deep convolutional networks for large-scale image recognition.
\newblock {\em CoRR}, abs/1409.1556, 2014.

\bibitem{tatler2007central}
B.~W. Tatler.
\newblock The central fixation bias in scene viewing: Selecting an optimal
  viewing position independently of motor biases and image feature
  distributions.
\newblock {\em Journal of vision}, 7(14):4--4, 2007.

\bibitem{rmac-ToliasSJ15}
G.~Tolias, R.~Sicre, and H.~J{\'e}gou.
\newblock Particular object retrieval with integral max-pooling of cnn
  activations.
\newblock In {\em International Conference on Learning Representations}, 2016.

\bibitem{walther2006modeling}
D.~Walther and C.~Koch.
\newblock Modeling attention to salient proto-objects.
\newblock {\em Neural networks}, 19(9):1395--1407, 2006.

\bibitem{Instre}
S.~Wang and S.~Jiang.
\newblock {INSTRE}: A new benchmark for instance-level object retrieval and
  recognition.
\newblock {\em ACM Trans. Multimedia Comput. Commun. Appl.}, 11(3):37:1--37:21,
  Feb. 2015.

\bibitem{Wu2017}
Y.~Wu, H.~Liu, J.~Yuan, and Q.~Zhang.
\newblock Is visual saliency useful for content-based image retrieval?
\newblock {\em Multimedia Tools and Applications}, Jul 2017.

\bibitem{wu2017visual}
Y.~Wu, H.~Liu, J.~Yuan, and Q.~Zhang.
\newblock Is visual saliency useful for content-based image retrieval?
\newblock {\em Multimedia Tools and Applications}, pages 1--24, 2017.

\bibitem{iSUN}
P.~Xu, K.~A. Ehinger, Y.~Zhang, A.~Finkelstein, S.~R. Kulkarni, and J.~Xiao.
\newblock Turkergaze: Crowdsourcing saliency with webcam based eye tracking.
\newblock {\em arXiv preprint arXiv:1504.06755}, 2015.

\bibitem{BabenkoL15}
A.~B. Yandex and V.~Lempitsky.
\newblock Aggregating local deep features for image retrieval.
\newblock In {\em 2015 IEEE International Conference on Computer Vision
  (ICCV)}, pages 1269--1277, Dec 2015.

\bibitem{yu2015multi}
F.~Yu and V.~Koltun.
\newblock Multi-scale context aggregation by dilated convolutions.
\newblock {\em arXiv preprint arXiv:1511.07122}, 2015.

\bibitem{deconvolution}
M.~D. Zeiler, G.~W. Taylor, and R.~Fergus.
\newblock Adaptive deconvolutional networks for mid and high level feature
  learning.
\newblock In {\em 2011 International Conference on Computer Vision}, pages
  2018--2025, Nov 2011.

\bibitem{zhang2013saliency}
J.~Zhang and S.~Sclaroff.
\newblock Saliency detection: A boolean map approach.
\newblock In {\em Proceedings of the IEEE international conference on computer
  vision}, pages 153--160, 2013.

\bibitem{LSUNchallenge}
Y.~Zhang, F.~Yu, S.~Song, P.~Xu, A.~Seff, and J.~Xiao.
\newblock Large-scale scene understanding challenge: Eye tracking saliency
  estimation.

\end{thebibliography}
}

\end{document}